# Macroscale Fracture Surface Segmentation via Semi-Supervised Learning Considering the Structural Similarity

Johannes Rosenberger [a] [b] [*], Johannes Tlatlik [b], Sebastian Münstermann [a]

[a] Steel Institute, RWTH Aachen University, Intzestraße 1, D-52072 Aachen, Germany

[b] Fraunhofer-Institute for Mechanics of Materials IWM, Wöhlerstraße 11, D-79108 Freiburg, Germany

[*] Corresponding author: johannes.rosenberger@iwm-extern.fraunhofer.de

## Abstract

To this date the safety assessment of materials, used for example in the nuclear power sector, commonly relies on a fracture mechanical analysis utilizing macroscopic concepts, where a global load quantity K or J is compared to the materials fracture toughness curve. Part of the experimental effort involved in these concepts is dedicated to the quantitative analysis of fracture surfaces. Within the scope of this study a methodology for the semi-supervised training of deep learning models for fracture surface segmentation on a macroscopic level was established. Therefore, three distinct and unique datasets were created to analyze the influence of structural similarity on the segmentation capability. The structural similarity differs due to the assessed materials and specimen, as well as imaging-induced variance due to fluctuations in image acquisition in different laboratories. The datasets correspond to typical isolated laboratory conditions, complex real-world circumstances, and a curated subset of the two. We implemented a weak-to-strong consistency regularization for semi-supervised learning. On the heterogeneous dataset we were able to train robust and well-generalizing models that learned feature representations from images across different domains without observing a significant drop in prediction quality. Furthermore, our approach reduced the number of labeled images required for training by a factor of 6. To demonstrate the success of our method and the benefit of our approach for the fracture mechanics assessment, we utilized the models for initial crack size measurements with the area average method. For the laboratory setting, the deep learning assisted measurements proved to have the same quality as manual measurements. For models trained on the heterogeneous dataset, very good measurement accuracies with mean deviations smaller than 1 % could be achieved, showcasing the enormous potential of our approach.

# Nomenclature

| | |
|---|---|
| $a_0$ | Initial crack size |
| $a_k$ | Starter notch length |
| $B$ | Specimen thickness |
| $B_N$ | Net specimen thickness |
| $c$ | Contrast |
| $Dice$ | Dice coefficient |
| $K$ | Stress intensity factor |
| $K_{Jc}$ | Elastic-plastic fracture toughness |
| $\mathcal{L}$ | Total loss |
| $\mathcal{L}_{s/u}$ | Supervised / unsupervised loss |
| $l$ | Luminance |
| $lr$ | Initial learning rate for model training |
| $n_{pixels}$ | Number of pixels per class in the ground truth mask |
| $n_{classes}$ | Number of classes in the ground truth mask |
| $N$ | Number of images |
| $p$ | Probability of augmentation application |
| $s$ | Structure |
| $W$ | Specimen width |
| $x$ | Model input |
| $y_{true/pred}$ | Ground truth / predicted labels |
| $z$ | Model output |
| $\Delta a$ | Mean absolute / relative measurement error |
| $\lambda$ | Consistency weight |
| $\mu_{u/l}$ | Ratio of unlabeled to labeled images for semi-supervised learning |
| $\mu_d$ | Pairwise difference of measurements |
| $\mu$ | Mean Value |
| $\sigma$ | Standard deviation |
| $\tau$ | confidence threshold |
| $\alpha, \beta, \gamma$ | weighting exponents |
| 5PA | 5-Point Average measurement |
| AA | Area Average measurement |
| CE | Cross Entropy loss |
| C(T) | Compact tension specimen |
| HAR | Harmonized dataset |
| HET | Heterogeneous dataset |
| HOM | Homogeneous dataset |
| mC(T) | Mini Compact Tension specimen |
| $mIoU$ | Mean Intersection over Union |
| SE(B) | Single Edge Notch Bending specimen |
| $SSIM$ | Structural Similarity Index Measurement |
| SSL | Semi-Supervised Learning |

# 1. Introduction

For the safety assessment of materials, used for example in the nuclear power sector, commonly a fracture mechanical analysis is performed. In practice this relies on macroscopic concepts where a global load parameter, such as the stress intensity factor K, is compared with the material's fracture toughness curve. For this purpose, a resource and time intensive manual classification of the present fracture mechanisms (brittle and/or ductile fracture, fatigue, etc.) is performed and the fracture surface must be measured (e.g., the initial crack size $a_0$). In accordance with the standards (e.g., ASTM [1–3]), the validity of the test has to be ensured. Additionally, the measurements usually require special equipment, such as light microscopes, and the image analysis has a subjective character. Considering the different settings in laboratories, the fracture mechanical analysis can be challenging even to a trained expert. To avoid a user-dependent analysis or even bias, and guarantee economic efficiency, comparability and, above all high quality of the evaluation across laboratories, image-based deep learning strategies can be implemented.

Contrary to classic image processing, used for example to automatically classify fracture surface morphologies [4, 5], deep learning approaches can effectively deal with perturbations like noise, contrast gradients, and other artifacts that are present in the data [6]. Famous semantic segmentation network architectures like U-Net [7] or Google's DeepLabv3+ [8] utilize encoder-decoder structures for feature map aggregation and restoration of the original resolution. This makes the models much more robust and allows for good generalization, which provides utility in, for example, automated measurement or detection of fracture mechanism features. In the context of materials science, state of the art deep learning approaches have mainly been used for quantitative microstructure analysis. Azimi et al. [9] used fully-convolutional neural networks (FCNNs) to classify low carbon steel microstructures. Recently, Ackermann et al. [10] showed the potential of multiple sequentially used models for the instance segmentation of martensite-austenite islands in different bainitic steels. In the fields of damage detection Thomas et al. [11] developed a deep learning framework for automated quantitative analyses of fatigue-induced surface damage. They investigated different augmentation strategies for material domain dependency in combination with the famous U-Net architecture [7]. Automated, and quantitative analysis of fracture mechanisms has been performed on a microscale level by Schmies et al. [12]. To support damage mechanism analysis in fractography they are developing a software based on deep learning with the goal of distinguishing and segmenting up to 18 different fracture mechanisms. To overcome objective bias, they also focused on training their models based on the labeling effort of 40 experts. Other microscale approaches [13, 14] had the goal of dimple detection for quantitative fractography utilizing the U-Net++ architecture [15]. Bastidas-Rodriguez et al. [16, 17] showed, that deep neural networks are suitable for the reliable classification of photo macrographs and SE images.

Although the utilization in microstructure and fractography analysis show the potential of deep learning segmentation approaches, the semantic segmentation of fracture surfaces on a macroscopic level poses a new field of application. Prior to this work we showed, that initial crack size measurements based on segmentation model outputs and the area average method are as precise as manual expert measurements [18]. While these models successfully deal with common perturbations of the SE(B)-specimens such as rust, color stains, and engravings (e.g., specimen numbers), out-of-domain generalization, for example to alternate imaging conditions in different laboratories was not given. This illustrates a knowledge gap, which this paper intends to close. This work will demonstrate how a suitable semantic segmentation framework is developed for macroscopic fracture surface images to gain advantages in experimental practice. Most importantly, it will not simply provide a suitable framework, but the intention is to analyze in depth the effect of types of images and advanced strategies regarding this specific use case. This is the main novelty of this paper, which is further highlighted by two main aspects. Firstly, the used data base is of unique quality, since it represents hundreds of fracture surfaces from different domains regarding materials, specimen types and laboratory conditions. Secondly, our framework gives insights on the use of advanced strategies to optimize the results such as semi-supervised learning (SSL) and augmentation pipelines for weak-to-strong consistency regularization in order to keep the required labeling effort as low as possible and improve efficiency of the model. This can be especially useful in practice. Finally, we were able to demonstrate the benefit for fracture mechanics assessment by using the trained and well-generalizing models for automated initial crack size measurements required for fracture toughness calculations.

## 2. Materials and methods

2.1 Data

The whole database includes 636 macroscale fracture surface images (top view) kindly provided from various international testing facilities from fracture mechanics testing (see Figure 1). These different subsets of images vary greatly in terms of material, toughness, specimen geometry, lighting conditions and image quality, which is the ideal basis for our framework. In order to specifically analyze how such a framework performs on different types of distinct data, the three investigated datasets, "homogeneous (HOM)", "heterogeneous (HET)", and "harmonized (HAR)" were established based on image similarity. Details on this distinction will be explained later. This procedure allows an evaluation of a model's performance with respect to its overall generalization. While the homogeneous dataset represents a typical isolated laboratory condition, the heterogeneous dataset can be considered as a complex real-world circumstance, while the harmonized dataset is a curated subset of the two.

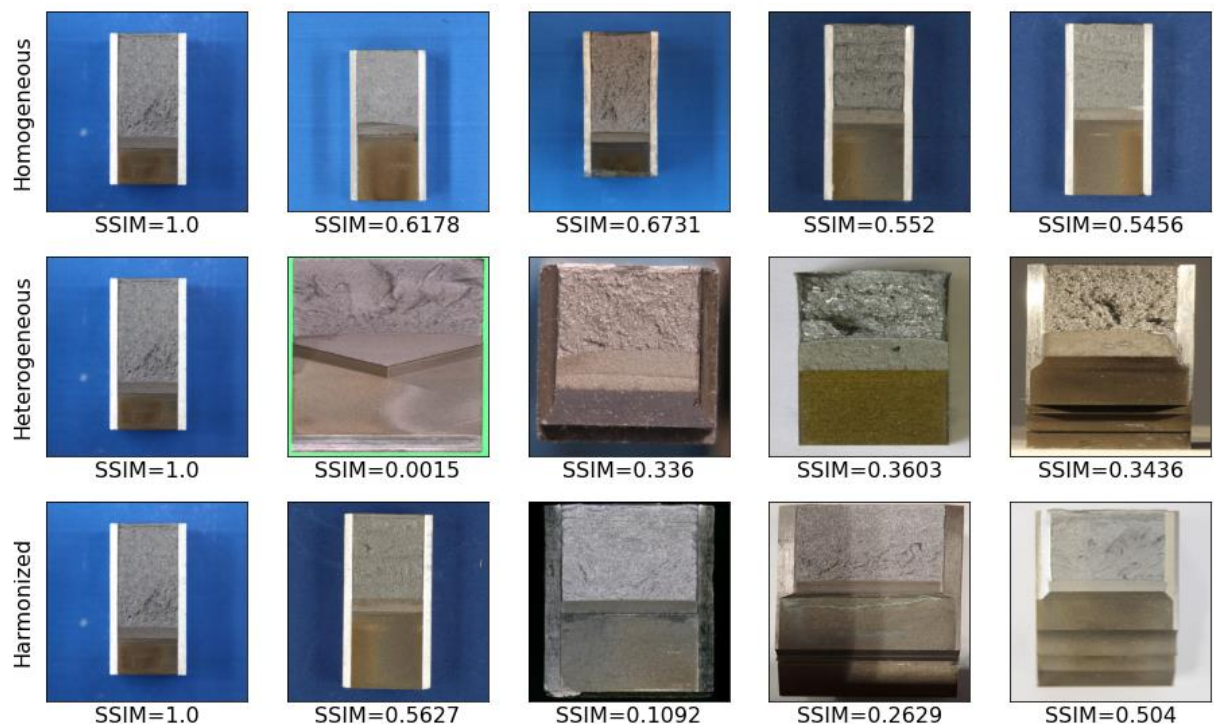

*Figure 1 Exemplary visualization of the dataset compositions and structural similarity of each image compared to the first image per row*

Semantic segmentation requires ground truth masks that are the equivalent to a pixel-precise classification of the input images. The images were labeled with the Python tool Labelme [19] suitable for polygonal annotation. To counter subjective bias, 6 experts participated in the labeling process, and labeled a total of 264 images. To guarantee sufficient labeling precision, the images were labeled at their original resolution and only then resized on the fly for training. The resulting ground truth masks have the same shape as the input image. In total 7 classes are distinguished: side groove, erosion notch, fatigue precrack, ductile fracture, brittle fracture, other, and background. The "other" class includes all parts of the specimen that were not explicitly mentioned in the list before e.g., gauge notches.

The datasets represent fracture surface images as they are photographed in daily use. Therefore, the images were taken in different environments, for example regarding the background or lighting conditions, and different camera setups (e.g., digital cameras, microscopes) were used. These differences can cause imaging-induced variance. The structural similarity index measurement (*SSIM*) is a metric used to measure the similarity of two images [20], where the similarity of two samples $x$ and $y$ is based on the comparison measurements of luminance ($l$), contrast ($c$) and structure ($s$). Usually, the weighting exponents $\alpha$, $\beta$, and $\gamma$ are set equal to 1.

$$SSIM(x,y) = l(x,y)^\alpha \cdot c(x,y)^\beta \cdot s(x,y)^\gamma \qquad (1)$$

Figure 1 shows an exemplary visualization of the three dataset compositions. For each image the structural similarity compared to the first image of each row is given. A clear difference between the training data is visible both qualitatively (visual) and quantitatively (SSIM-values). The images for the homogeneous dataset are all taken in front of a blue background and result from previous projects [21–24] at the same laboratory. The other datasets additionally include the images from the other laboratories. Here not only the capture environment and setup differ, but also the specimen types and materials cover a wider range, as for example Chevron notched specimens are introduced in the heterogeneous dataset. For the homogeneous and the harmonized dataset different forms of imaging-induced variance, such as shadowing, lighting, and backgrounds, can be found in Figure 1.

*Table 1 Feature summary for the three investigated datasets and their structural similarity.*

| Feature / Dataset | Homogeneous (HOM) | Heterogenous (HET) | Harmonized (HAR) |
|---|---|---|---|
| $N$ | 200 | 588 | 314 |
| $N_{labeled}$ / $N_{unlabeled}$ ($\mu_{u/l}$) | 32 / 168 (5.25) | 96 / 492 (5.13) | 46 / 268 (5.83) |
| $N_{train}$ / $N_{val}$ | 24 / 8 | 72 / 24 | 30 / 16 |
| $N_{test}$ | 38 | 24 | 17 |
| Specimen types | SE(B)40x20 ($a_0/W$ = 0.5 and 0.3) | various SE(B), various (mini) C(T), Chevron-notched | various SE(B), various C(T) |
| Materials considered (extract) | 22NiMoCr3-7 + weld material | 22NiMoCr3-7 + weld material, 34CrNiMo6, 316L welds, S690QL, others | 22NiMoCr3-7 + weld material, 34CrNiMo6, S690QL |
| $\mu_{SSIM,train+val}$ / $\mu_{SSIM,test}$ | 0.560 / 0.556 | 0.370 / 0.356 | 0.459 / 0.510 |
| $\sigma_{SSIM,train+val}$ / $\sigma_{SSIM,test}$ | 0.031 / 0.024 | 0.117 / 0.115 | 0.113 / 0.073 |

Table 1 summarizes the composition and characteristics of the investigated datasets. The homogeneous dataset is the smallest dataset with 200 images total, which is still very extensive. It only contains images of SE(B)40x20 specimens of 22NiMoCr3-7, a German reactor pressure vessel steel, and its weld material with different starter notch lengths $a_k$ (crack depth ratio $a_0/W$ = 0.5 and 0.3) in front of a blue background. For the homogenous dataset all images were already available labeled. To analyze the influence of image similarity, as well as the ratio of unlabeled to labeled data $\mu_{u/l}$ on the training, and ensure comparability to the other datasets, only 32 labeled images were used, the other 168 images were used in the unsupervised learning pipeline during training. The heterogeneous dataset is set up to train a segmentation model, capable of generalizing over a wide range of materials, specimen types and image environments. It consists of all 588 images with a large amount of different specimen types and materials, and backgrounds, of which 96 images are labeled (24 HOM ground truth masks). It covers fracture surfaces from all 6 participating laboratories. The harmonized dataset was introduced as a compromise between the homogeneous and the heterogeneous dataset and is a subset of the latter. It covers less specimen types and materials, but the ratio of unlabeled to labeled images $\mu_{u/l}$ is roughly equivalent to the distribution in the harmonized and homogeneous dataset. To gain true information on the generalization capability and avoid a bias of one image type in the labeled training data, not all labeled images that are used in the homogeneous dataset are used as labeled data in the semi-supervised datasets. These unused images rather serve as unlabeled training data. Furthermore, we investigated the effect of a stratified vs. a randomized approach for the generation of the validation dataset. For the stratified approach, the images were manually subdivided to avoid possible biased splits e.g., overrepresentation of a specimen type in validation data, which can occur with randomized approaches in datasets with many representations compared to the dataset size.

As desired, the datasets show clear differences in their structural similarity, when calculated with the exponents $\alpha$, $\beta$, and $\gamma$ set equal to 1. While the homogeneous dataset shows a mean structural similarity of roughly 56 %, the heterogeneous and the harmonized datasets are less similar. Also, the standard deviation is significantly higher for these datasets, underlining the clear difference to the homogeneous data. The test datasets show SSIM values and standard deviations in the same regime as their training and validation datasets. This ensures test data originates from the same distribution, which is crucial for deep learning approaches, and that the test datasets are suitable representatives for the model evaluation. In Figure 2 these differences are visualized in SSIM matrices for each dataset, respectively the test datasets.

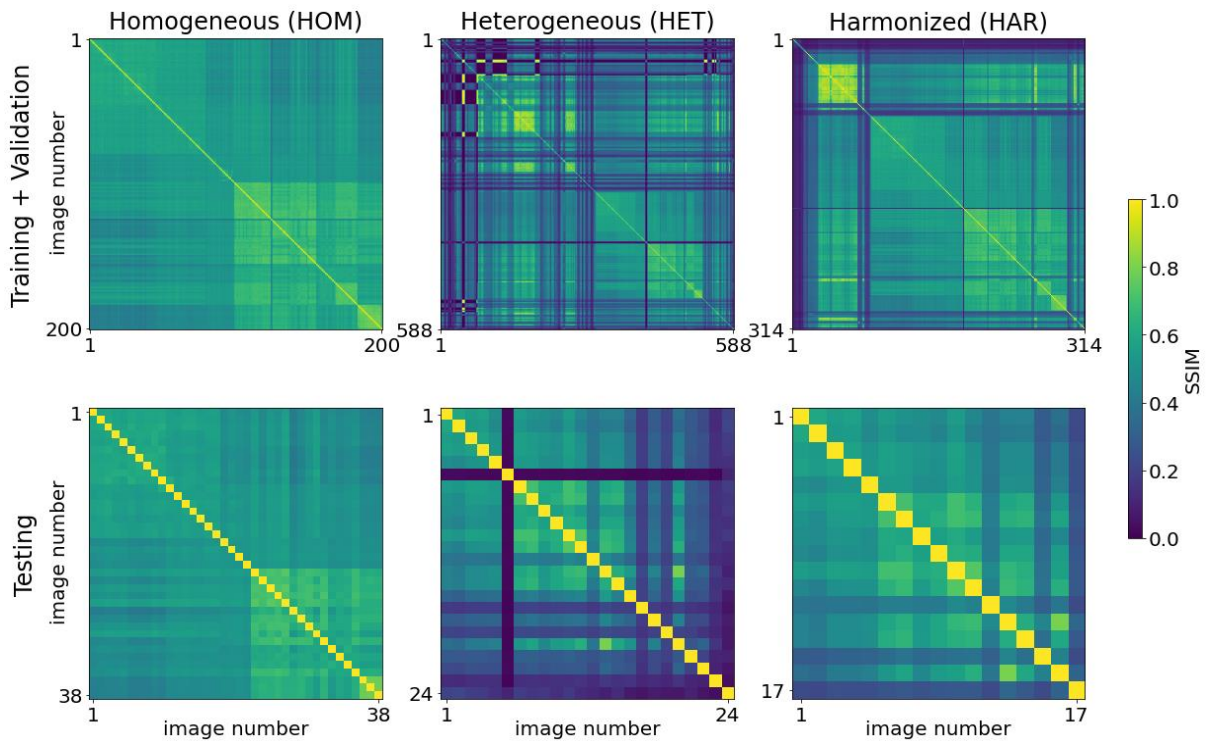

*Figure 2 Visualization of the SSIM analysis for the three investigated datasets (top: training and validation, bottom: test data)*

2.2 Pre-processing and augmentations
The size of the training images ranges from 420x169 to 6000x4000 pixels. In preparation of the training process all images are sliced into 4 equally sized patches, which are then resized to 512x512 pixels to serve as input for the segmentation models. By this, graphic processing unit (GPU) memory limitations are circumvented. For inference and testing, the patches are stitched back together, increasing the resolution of the prediction masks, which is desired side-effect for the measurement task.

The appearance of the fracture surface can differ due to predefined fixed parameters such as the material under consideration, different loading conditions respectively testing techniques, or the equipment used for capturing the image. Additionally, there are variables that can be influenced during image capturing e.g., lighting conditions, backgrounds, that will differ from lab to lab and even day to day. To increase the robustness, generalizability, and to avoid overfitting, augmentations were applied during training. Augmentations aim to mimic the occurring variance by applying various transformations to the input images and ground truth masks [25].

*Table 2 Description of individual augmentations implemented.*

| Augmentation Type | Description | Parameter(s) | $x_{lim}$ | $p$ |
|---|---|---|---|---|
| *Non-destructive transformations (used in weak augmentation pipeline)* | | | | |
| Affine transformations* | Linear shifting, scaling, and rotation | Shift, scale, and rotation limit | 0.0625, 0.1, 45 | 0.25 |
| Rotation & flip | Random 90° rotation and flipping | - | - | 1.0 |
| *Non-rigid transformations (used in weak augmentation pipeline)* | | | | |
| Grid distortion* | | Grid cells, distort limit | 5, 0.3 | 0.5 |
| *Non-spatial transformations (used in strong augmentation pipeline)* | | | | |
| Sharpening* | - | Alpha, lightness | (0.2, 0.5), (0.5, 1.0) | 0.25 |
| Blurring* | - | Blur limit | 7 | 0.2 |
| Channel shuffling* | Randomly rearrange RGB channels | - | - | 1.0 |
| Gaussian noise* | Add gaussian noise to input image | Loc | 0.05 | 0.1 |
| Brightness & contrast | Randomly change brightness and contrast | Brightness limit, contrast limit | 0.2, 0.2 | 1.0 |

*Using the Albumentations framework [26].

In Table 2, a description of the employed augmentation types is given. The augmentations are applied to the images with the specified limitations and probabilities on-the-fly prior to each epoch during training. The non-destructive and non-rigid transformations are applied to both, the input image, and the ground truth mask. The non-spatial transformations are only subject to the input images since they don't alter the information location but representation. Other non-rigid transformations like elastic transformations or strong optical distortions have not been considered because they don't resemble typical specimen representations found in real world settings. The specimen edges for example are typically straight and not wavy as after elastic transformations have been applied. Additionally, our approach requires the distinction between weak and strong augmentations, which will be considered in detail in the next section.

## 2.3 Network and Training Strategy

In computer vision, semantic segmentation is a task in which the goal is to categorize each pixel in an image. The architecture used is a DeepLabv3+ multiclass segmentation model as proposed by Chen et al [8], implemented in PyTorch, and illustrated in Figure 3 combined with a ResNet50 [27] backbone. In previous studies [18] on the homogeneous dataset with a supervised-learning approach, this architecture achieved the highest mean Intersection over Union (mIoU) scores of all tested network-backbone combinations. The mIoU denotes the mean of all class-wise Intersection over Union (IoU) values, where the class-wise area of intersection or overlap between the ground truth mask *A* and prediction mask *B* divided by the area of union between the prediction mask and the ground truth mask.

$$IoU = J(A,B) = \frac{|A \cap B|}{|A \cup B|} \tag{2}$$

$$mIoU = \frac{1}{n_{classes}} \sum_{j=1}^{n_{classes}} IoU_j \tag{3}$$

The network utilizes an encoder-decoder structure as it is known from other segmentation networks, such as the famous U-Net [7], U-Net++ [15] ,or DeepLabv3+ [8]. Typically, the encoder path captures higher semantic information by gradually reducing the feature map size, aggregating the information density. The decoder gradually recovers the spatial information lost during encoding with the help of skip connections. The result is a pixel-wise classification of the input image with the same spatial resolution as the input image.

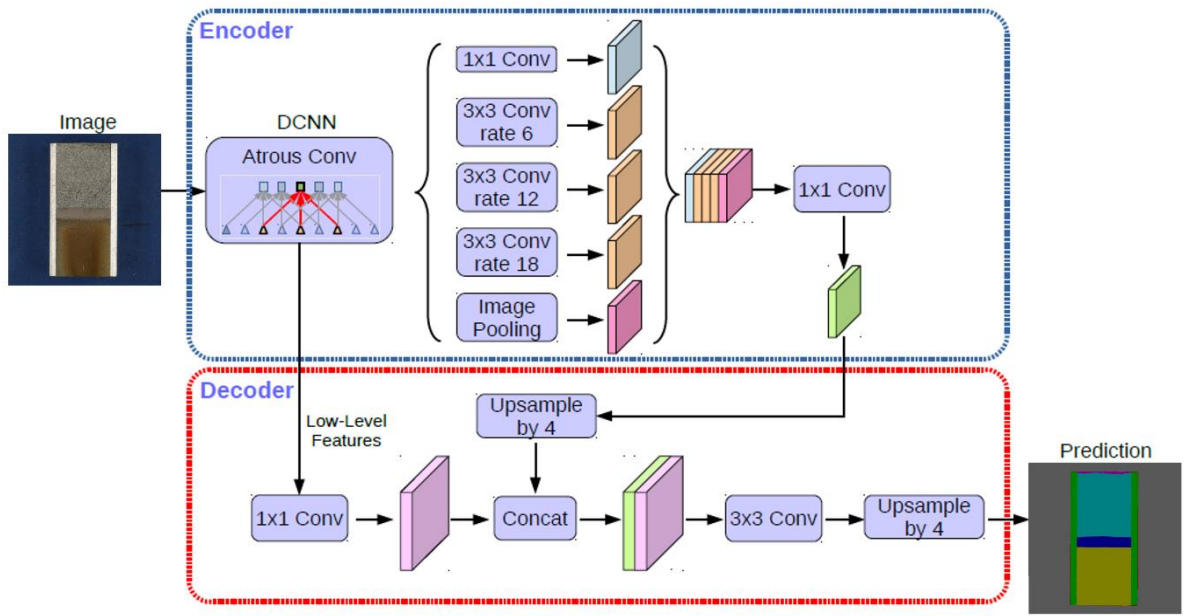

*Figure 3 DeepLabv3+ architecture as proposed by Chen et al [8]*

Semi-supervised learning strategies provide the ability to additionally train models with unlabeled data. This significantly reduces the amount of labeling work required because not all images have to be manually annotated, enabling the utilization of larger dataset sizes. Our approach to train the DeepLabv3+ models with unlabeled data is an adaptation of the FixMatch algorithm [28] for semantic segmentation. This so called weak-to-strong consistency regularization requires the implementation of two augmentation pipelines (weak and strong, see Table 2). With appropriate image-level augmentations this method has shown to achieve state-of-the-art generalizability [29].

The loss function $\mathcal{L}$ consists of a supervised loss $\mathcal{L}_s$ and an unsupervised loss $\mathcal{L}_u$ weighted with a consistency weight $\lambda$. The supervised loss is calculated based on the comparison of the labeled input $x$ and the model output $y$ via the sum of cross entropy (CE) loss (5) and the comparison of ground truth ($y_{true}$) and predicted labels ($y_{pred}$) via Dice loss (6). To prevent overfitting, the input faces the weak augmentations introduced earlier.

$$\mathcal{L} = \mathcal{L}_s + \lambda \mathcal{L}_u \tag{4}$$

$$CE(x,y) = -\sum_{j}^{n_{classes}} y_j \log \frac{e^{x_j}}{\sum_{i}^{n_{classes}} e^{x_i}} \tag{5}$$

$$Dice\ loss = 1 - Dice = 1 - \frac{2\sum_{pixels} y_{true} y_{pred}}{\sum_{pixels} y_{true}^2 + \sum_{pixels} y_{pred}^2} \qquad (6)$$

For the calculation of the unsupervised loss, the model is presented with a weak and a strong augmented version of the unlabeled image. For the strong augmentation pipeline, additional non-spatial transformations are employed. For both, the weak and the strong augmented images, pseudo-labels are generated. If the classification for a pixel in the weak transformation pipeline surpasses a confidence threshold $\tau$ of 0.8 both pseudo labels can be compared for this pixel. This prevents weak predictions from influencing the results especially in early stages of the training. The prediction confidence of each pixel for class $i$ can be calculated by applying a softmax activation function to the final layer (output) $z = y_{pred}$ of the neural network.

$$Softmax(z_i) = \frac{e^{z_i}}{\sum_j^{n_{classes}} e^{z_j}} \qquad (7)$$

A consistency loss can be calculated (consistency regularization), based on the differences between the pseudo-labels of the weakly perturbed image (acts as ground truth) and the strongly perturbed version. Additionally to the cross entropy loss and Dice loss, a complementary loss that compares the least likely predictions of the weak and strong augmented images (negative learning [30]) is used. Conveniently, the CE can once more be used for this purpose. The consistency weight is a time-dependent parameter and can be described as a sigmoid ramp-up [31] that reaches its limit after 200 epochs of training.

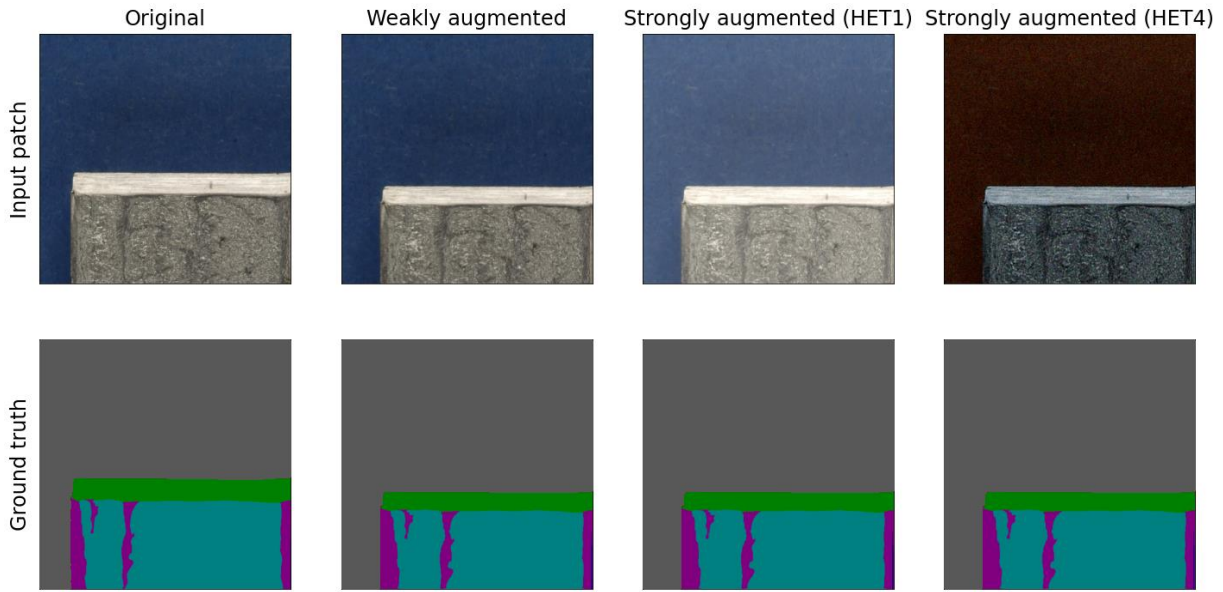

*Figure 4 Exemplary visualization of the outcome for the weak and strong augmentations of models HET1 and HET4 (top: input images, bottom: ground truth masks)*

In Figure 4, an example for a possible output of the weak and strong augmentation pipelines is given for two different strong augmentation strategies. The left column shows the input image and ground truth mask prior to any augmentation. For the weakly augmented pipeline, the input image, and its mask showcase non-destructive e.g., linear, and non-rigid transformations, such as grid distortion. In Figure 4, the first strongly augmented input image is additionally affected by brightness and contrast transforms, whereas the second is also affected by gaussian noise, blurring, and a channel shuffling (each with a probability $p$, as seen in Table 2). The ground truth masks are not affected by these augmentations. Therefore, no manual labeling by an expert is required for these images. The model is apparently able to overcome the difficulties of predicting the strong augmented image and generalizes the better, the closer the prediction masks are both images. The closer the predictions are, the lower is the consistency loss. Therefore, great emphasis was put on finding the right augmentation strategy for the weak-to-strong consistency regularization in this study.

The implemented code is publicly available online [32], and was partially adapted for this study. Specifically, the dataloader was customized to ensure compatibility with our used image format and the tailored augmentation pipelines implemented with Albumentations [26]. To enable comparability of the validation losses between the trainings, the Dice loss function [33] tested in a previous study [18] was implemented, and the originally used Hausdorff distance was discarded as it can cause errors [34]. To accelerate the training, a ResNet50 backbone, that was pretrained with 7 classes on ImageNet [35], was employed from the GitHub repository "Segmentation Models Pytorch" [36]. The trainings were performed on a NVIDIA V100S GPU, where training a single model took roughly 11 to 13 hours. In Table 3, an overview of the trained models and their differences regarding validation data selection and training strategies is given in accordance with the methods described prior. For the training of reference models, the oracle models REF1-3 were trained fully-supervised with the complete training data.

*Table 3 Model and training strategy summary*

| Model | Dataset | Validation data | Training strategy | Comment |
|---|---|---|---|---|
| REF1 | HOM | Randomized | Supervised | - |
| REF2 | HET | Randomized | Supervised | - |
| REF3 | HET | Stratified | Supervised | - |
| HET0-8 | HET | Stratified | Semi-supervised | Numbers denote different augmentation strategies. |
| HOM1/4 | HOM | Randomized | Semi-supervised | |
| HAR1/4 | HAR | Stratified | Semi-supervised | |
| MAX1/4 | HET | Stratified | Semi-supervised | All additional HOM labels available were added to the supervised pipeline. |

# 3. Results and Discussion

The investigations on a semi-supervised training framework for macroscale fracture surface segmentation were carried out on the heterogeneous dataset. First, reference models were trained in a fully-supervised fashion to determine the validation data selection process. In the next step, the augmentation strategies for the weak-to-strong consistency regularization pipeline were tested. Here, 9 different strategies were tested on the heterogeneous dataset and analyzed for their strengths and weaknesses. Additionally, the influence of the number of provided labeled images for training on the prediction quality of the models has been investigated. To conclude the investigations on the model prediction quality and generalizability, the influence of the structural similarity of the provided training data was tested with the setting determined in the steps before. The homogeneous and harmonized datasets were used for this purpose. With the highest scoring model for each dataset, the measurement precision based on the prediction masks was tested. Table 4 lists the obtained prediction results for all used augmentation approaches with respect to the used data.

*Table 4 Segmentation results on the test data for the models trained with different augmentation pipeline configurations on reference and heterogeneous data (+ augmentation applied, - augmentation not applied)*

| Model | Weak augmentation pipeline | | | | Strong augmentation pipeline | | | Metrics | | | | | | |
|---|---|---|---|---|---|---|---|---|---|---|---|---|---|---|
| | Affine transformations | Rotation & flip | Grid distortion | Brightness & contrast | Gaussian noise | Channel shuffling | Sharpening & Blurring | $IoU_{side\ groove}$ [%] | $IoU_{erosion\ notch}$ [%] | $IoU_{fatigue\ precrack}$ [%] | $IoU_{ductile\ fracture}$ [%] | $IoU_{brittle\ fracture}$ [%] | $IoU_{other}$ [%] | mIoU [%] |
| REF1* | - | + | - | + | - | - | - | 97.3 | 99.4 | 98.0 | 73.0 | 98.5 | - | 94.8 |
| REF2* | - | + | - | + | - | - | - | 78.4 | 89.4 | 74.1 | 25.9 | 84.9 | 21.0 | 67.3 |
| REF3* | - | + | - | + | - | - | - | 79.9 | 91.1 | 77.4 | 36.3 | 91.1 | 57.9 | 78.7 |
| HET0 | - | + | - | + | - | - | - | 86.5 | 94.8 | 89.7 | 59.4 | 91.3 | 47.4 | 84.6 |
| **HET1** | + | + | + | + | - | - | - | **91.0** | **97.8** | **93.3** | **58.4** | **90.6** | **92.5** | **88.8** |
| HET2 | + | + | + | + | + | - | - | 90.3 | 93.9 | 92.2 | 54.3 | 89.6 | 63.9 | 85.3 |
| HET3 | + | + | + | + | + | + | - | 86.6 | 95.1 | 93.7 | 54.6 | 89.9 | 70.6 | 86.0 |
| **HET4** | + | + | + | + | + | + | + | **90.2** | **97.9** | **92.2** | **54.5** | **89.7** | **92.3** | **87.7** |
| HET5 | + | + | + | + | - | + | + | 90.5 | 97.5 | 91.8 | 54.2 | 90.0 | 91.2 | 87.5 |
| HET6 | + | + | + | + | - | - | + | 89.9 | 91.6 | 91.4 | 55.0 | 89.5 | 60.6 | 84.8 |
| HET7 | + | + | + | + | + | - | + | 90.5 | 93.6 | 93.7 | 55.3 | 90.0 | 70.4 | 86.6 |
| HET8 | + | + | + | + | - | + | - | 90.5 | 98.0 | 94.1 | 53.7 | 90.2 | 80.5 | 87.4 |

*Trained fully-supervised.

It can be found that for the semi-supervised FixMatch [28, 32] approach, the training was successful as all models trained on the heterogeneous dataset surpass a mIoU of 84.5 %.

3.1 Validation data selection

The prediction quality was evaluated for each model by calculating the average mIoU score for all images in the test dataset. In Table 4 the results for the different augmentation pipeline configurations are given. Additionally, the class-wise Intersection over Union (IoU) values are given for all classes but the background class as it is greater or equal to 97.8 % for all models. Note, that for the calculation of the mIoU the background class was considered. As a benchmark, three models were trained in a fully-supervised fashion, where a simple augmentation strategy, utilizing only rotation and flips, as well as brightness and contrast transformations, was used. REF1 was trained on the homogeneous dataset (200 labeled images), whereas REF2 and REF3 were trained on the heterogeneous dataset with only 96 labeled images. The latter models were used to study the effects of a stratified versus a randomized validation data selection approach. Both models were trained with the same augmentation strategy on the heterogeneous dataset. While REF2 utilized a randomized approach for validation data selection, REF3 was trained with manually selected validation images. The goal of this approach is to avoid the possibility of a biased training-validation-split by manually distributing the data. Utilizing the stratified approach, the segmentation performance was increased by roughly 11 %. For a real-world dataset with many data representations as the heterogeneous dataset, the stratified approach is more suitable than the randomized one. In accordance with the structural similarity and the number of available labeled training data, the mIoUs achieved on the homogeneous dataset were much higher than for the models trained on the heterogeneous data. This is as expected. However, the results for the REF1 model (randomized approach) show the training success was not dependent on a stratified validation data selection as a mIoU of 94.8 % was achieved. The images in this dataset are very similar due to the laboratory setting. Therefore, the randomized validation data selection will always be representative of the whole dataset. Since the goal was to train a well-generalizing, domain-independent model on the heterogeneous dataset, where randomness induced variance due to data selection would not be beneficial, the further investigations were performed with the stratified approach.

### 3.2 Weak and strong augmentation strategies

The investigations on the augmentation pipelines are denoted by a number for the different augmentation strategies (HET0-8), and the results are given in Table 4. First, the weak augmentation strategy was determined by training two models, where the strong augmentation pipeline was kept constant and simple (HET0/1). The approach of maximizing the number of possible transformations (HET1), rather than only applying very simple transformations, such as "rotation & flip", and "brightness & contrast" (HET0), benefitted the prediction quality (+4.2 %). Therefore, the strong augmentation strategy was determined with these settings for the weak augmentation pipeline. For the strong augmentation pipeline the simplest strong augmentation strategy, denoted HET1, showed the best prediction performance. Here the input images were only affected by one additional perturbation, namely a random brightness and contrast transformation. Just like for the weak augmentations the maximization of possible transformations (HET4) also achieved a high overlap between ground truth and prediction masks. Strong augmentation strategies, where the channel shuffle transformation was added (HET5/8) showed better results than pipelines with implemented gaussian noise (HET2/3/6/7). Arguably, the gaussian noise poses too much of a perturbation to the models since it clearly changes the texture and structure of the input image (see Figure 4). It may have forced the models to learn features that are not part of the domain, and important features for pixel-classification could be lost. The channel shuffle however did not affect the texture of the images, so no information was compromised. For the side groove, erosion notch, fatigue precrack, and brittle fracture class all HET models showed good prediction qualities with IoU scores of 86.5 % or higher. HET1 and HET4 showed the highest IoUs for these classes. For the ductile fracture class, the models trained with the minimalistic strong augmentation pipeline achieved better results than the maximized strategy (roughly +4 %). The model prediction qualities for the "other" class varied a lot across the different augmentation pipelines (IoUs 47.4-92.5 %), but again the minimalistic and the maximalist approach achieved the best results. In general, the better overall prediction quality for HET1 and HET4 can be explained by their capability to better recognize the "other" class pixels. On top of that HET1 performs better on the ductile fracture class, ranking it in the first place of the investigated strong augmentation pipelines.

Comparing the results qualitatively, Figure 5 shows that the semi-supervised learning approach increased the model prediction quality for all specimen types regarding the heterogeneous dataset. For all images, the specimen edges are predicted more distinct by the SSL models. While for the reference model REF3 (trained fully-supervised), all but the SE(B)40x20 specimens (a, b) showed low mIoU scores, the HET4 model only performed bad on the mini-C(T), denoted as mC(T), specimen with a thickness $B$ of 1.62 mm (d). The result ranks even below the reference model, as the fatigue precrack and the ductile fracture are heavily misclassified as brittle fracture. Nevertheless, an improvement was made for the recognition of the "fatigue precrack" crack front curvature, which is not indicated at all for REF3. The HET0 model outperformed the reference model for every image shown. For the mC(T) specimen with thickness 4.17 mm (e) and the C(T) specimen (f) an increase in prediction quality of roughly 21 and 18 % can be found. Small misclassifications are visible for the fatigue precrack of (c) and (f). An overprediction is recognizable for the Chevron notched specimen. The prediction visually matches the ground truth mask very well (especially HET4), while the mIoU is at roughly 80 % only. The reason for that are a few "brittle" pixels, even barely visible in full scale, that were misclassified as ductile fracture with no ductile fracture present in the image. If a model predicts a class in an image where this class is not present, the IoU for this specific class is set to 0, decreasing the images mIoU.

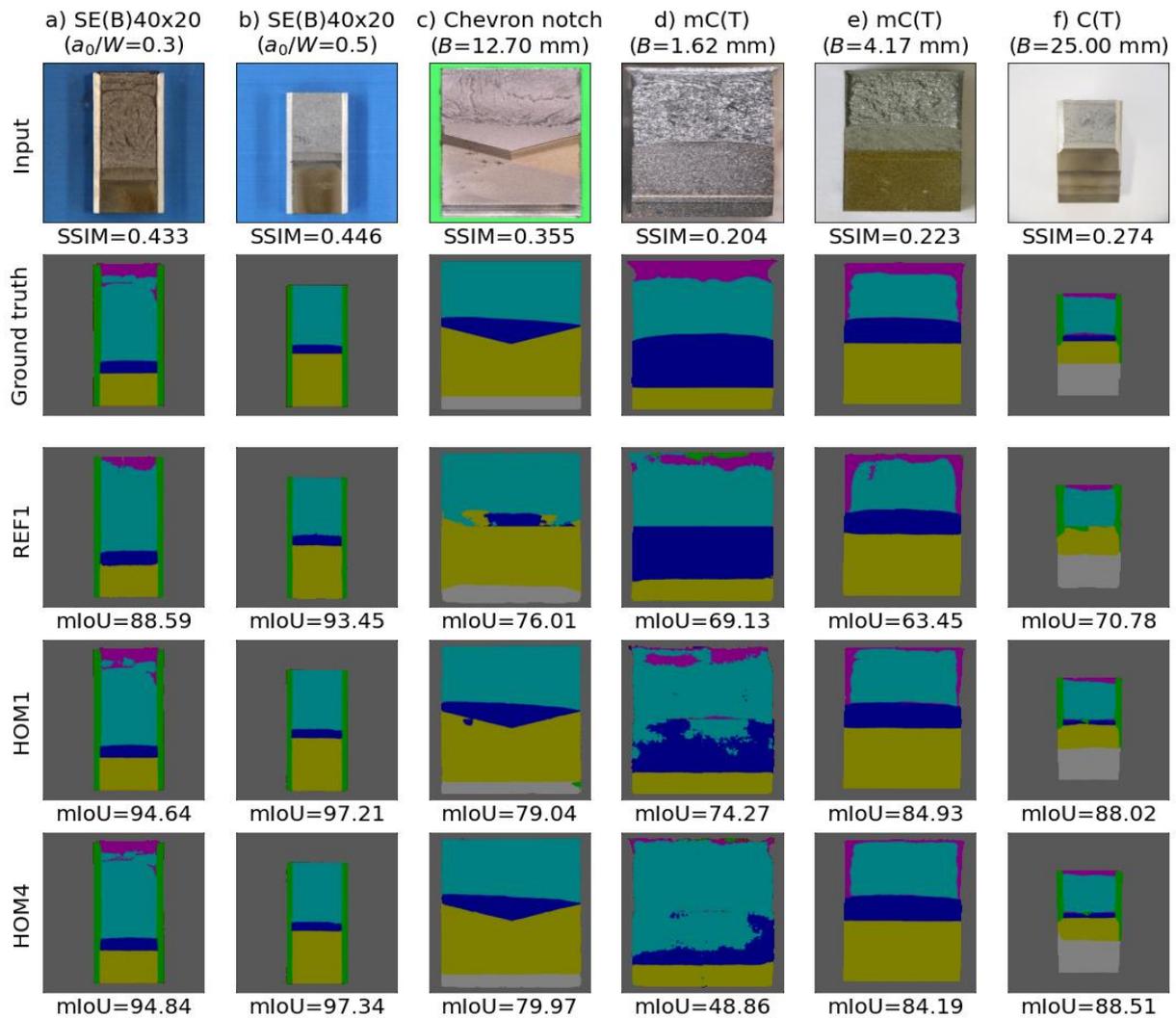

*Figure 5 Comparison of prediction quality for selected images from the heterogeneous test dataset*

In Figure 6 the mIoU scores for each image in the test dataset are plotted to the number of pixels per class $n_{pixels}$ for HET1 and HET4. Note that due to the logarithmic scale, $n_{pixels}=10^0$ stands for images, where the specific class is not present. The HET4 outlier present for all classes is the mC(T) specimen (d) shown in Figure 5. For the side groove, the ductile fracture, and the "other" class images exist, where these classes are not present, and a misclassification of a single pixel as these classes leads to IoUs of 0. These images show mostly Chevron notched images, and specimens without side grooves or images, where no gauge notch can be found (e.g., mC(T) specimens). On the other hand, the erosion notch class, the fatigue precrack, and brittle fracture occur in every image. The described misclassification effect occurred less often for the HET1 model. Therefore, the minimized strong augmentation pipeline helped the model to learn to not misclassify pixels in images where a specific class is not present better than the maximized augmentation pipeline. Overall, these results also clarify that the semi-supervised approach is the only viable option for training a well-generalizing and robust model, that can cover various types of specimens and laboratory settings and configurations if only a limited amount of labeled data is given.

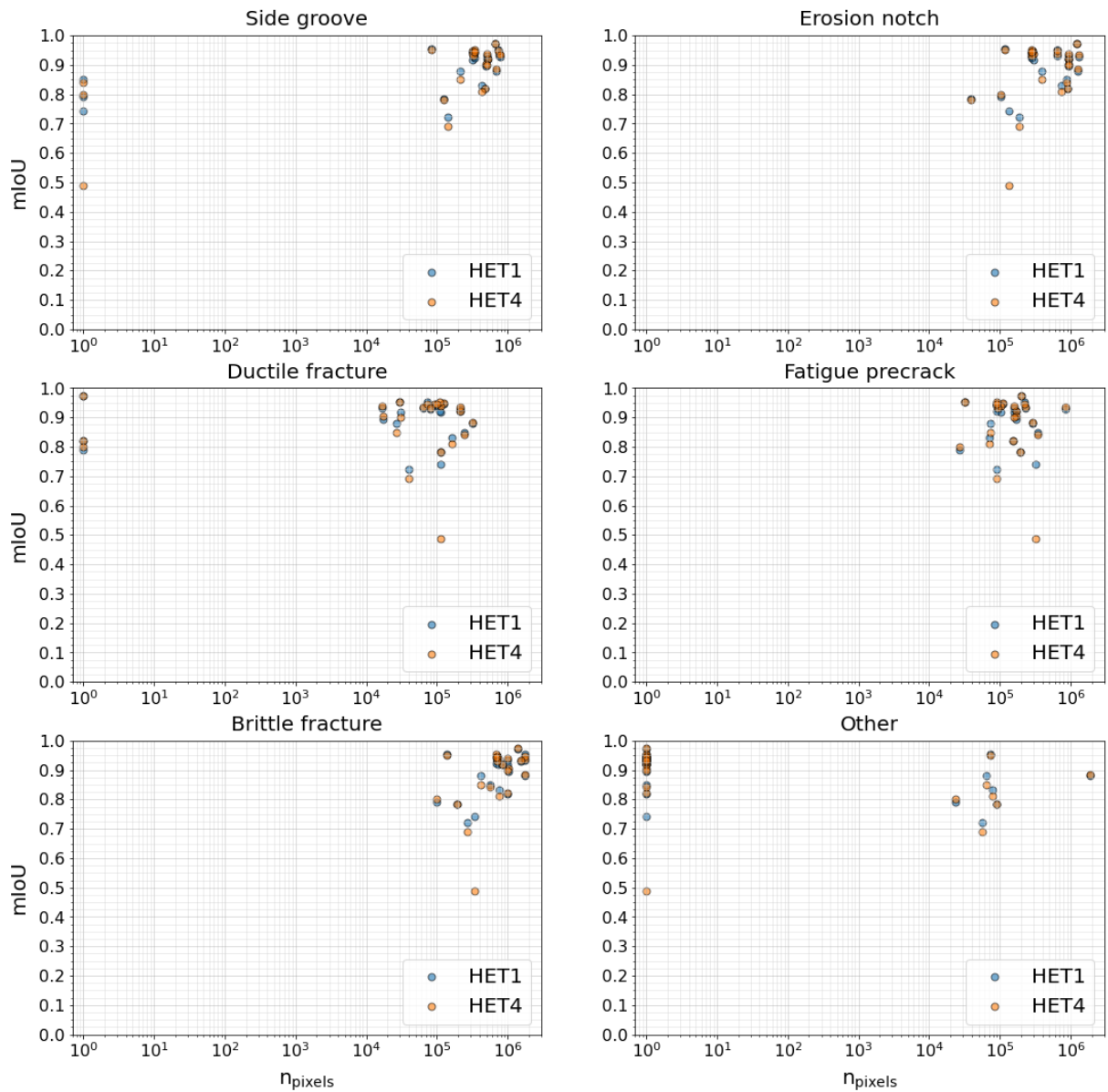

*Figure 6 Segmentation results in dependence of $n_{pixels}$ for the HET1 and HET4 model ($n_{pixels}=10^0$ denotes images, where the specific class is not present)*

In Figure 7 further analysis of the mIoU scores with respect to the number of classes $n_{classes}$ in the ground truth masks respectively in the input images has been performed. For each augmentation strategy, the standard deviations are indicated by the black error bars, and the orange shading indicates the minimum and maximum values. In general, the variance for the 5 class images is much higher than for the other numbers of classes. Comparing HET0 to the other models it could be found that the change of the weak augmentation pipeline (maximizing the number of possible transformations) helped to significantly raise the average mIoU and reduce the standard deviation for fracture surface images where only 5 classes are present. The HET1 model showed the biggest standard deviation reduction for these images, which correlates with the observation made for the missing class IoUs above (see Figure 6), where the misclassification effect occurred less often. This means that the number of misclassifications as non-present classes is strongly reduced if the complexity of the weak augmentations is increased. For the ground truth masks with 6 classes present (e.g., the SE(B)40x20 specimens in Figure 5) the highest mIoU scores were achieved. For HET1 and HET4, as well as other good performing models (HET5/8), the standard deviation is strongly reduced, whereas the other augmentations strategies do not seem to influence the prediction quality for these images. The augmentation strategies 4, 5 and 8 have the channel shuffling transformation in common which seems to benefit the prediction quality for these image types. However, for the HET1 and HET4 model this effect is the most pronounced. The results for the 7 class images were not affected too much by the different augmentation strategies, as the mean value and standard deviation did not change significantly. This implies that the models did not benefit from a particular strong augmentation pipeline to generalize over the different representations. Furthermore, it can be deduced from this that in images showing the maximum number of classes no classes are forgotten i.e., not detected at all.

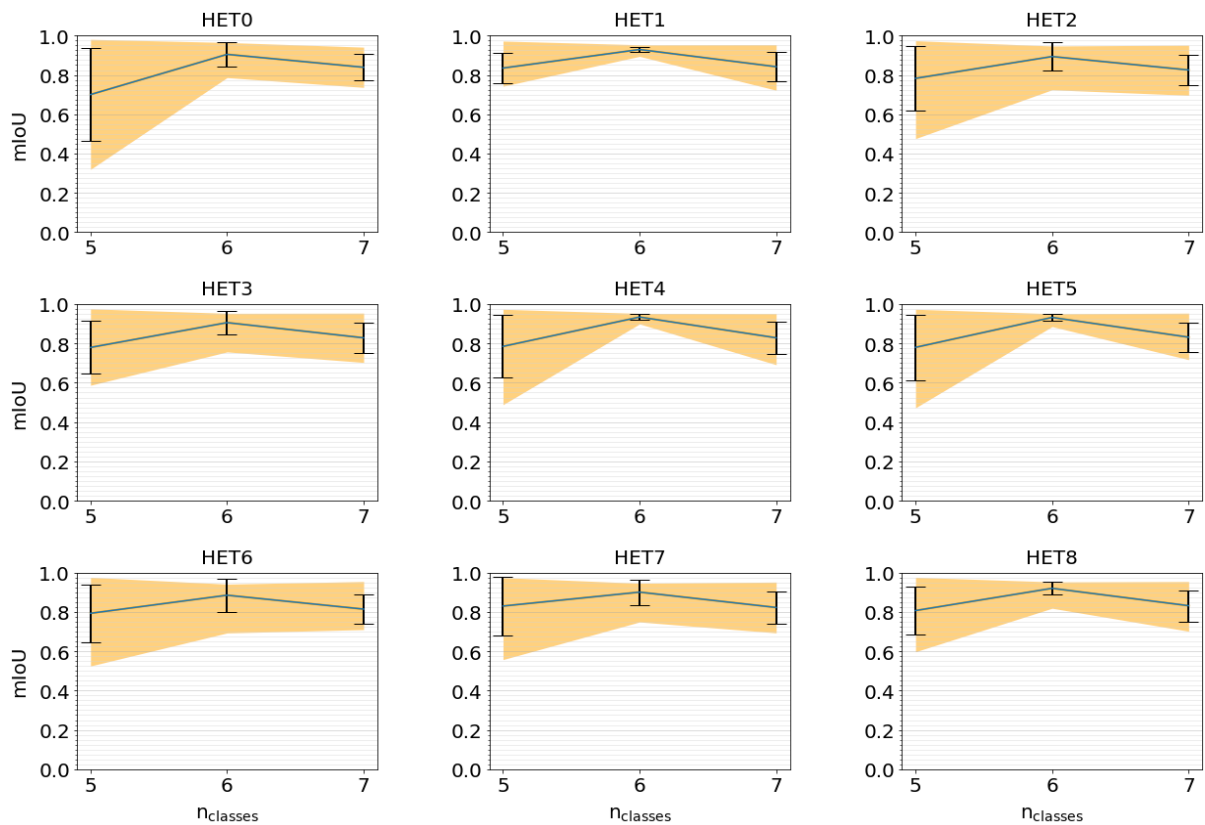

*Figure 7 Segmentation results in dependence of $n_{classes}$ (in ground truth masks) for the different augmentation pipeline configurations*

## 3.3 Structural similarity and generalizability

In this chapter the influence of the structural similarity on the training success and model prediction and generalization capabilities is described. Based on the results described above, the following investigations on the influence of the structural similarity were carried out on models trained with the augmentation pipelines used in HET1 and HET4 (see Table 5). The introduced models are therefore denoted with the same numbers for the applied augmentation strategy. For the homogeneous dataset the models were trained with a 7-class backbone although only 6 classes are present in this dataset. As the results for HOM4 show, this did not pose a problem for the semi-supervised learning approach. The model learned that there is no "other" class in the training data and no misclassifications could be observed for this class. However, the HOM1 model showed the lowest overall mIoU score, behaving contrary to the findings shown before, where the minimalistic strong augmentation approach was the most precise. The prediction quality of HOM4 is very good throughout all classes but the HOM1 model cannot properly detect fatigue precrack (mIoU of 12.1 %) and ductile fracture (30.1 %). Hence, the model is not utilizable for any kind of measurement related task.

*Table 5 Segmentation results on the test data for the models trained with augmentation pipeline configurations 1 and 4 on all datasets (+ augmentation applied, - augmentation not applied)*

| Model | Weak augmentation pipeline | | | | Strong augmentation pipeline | | | Metrics | | | | | | |
|---|---|---|---|---|---|---|---|---|---|---|---|---|---|---|
| | Affine transformations | Rotation & flip | Grid distortion | Brightness & contrast | Gaussian noise | Channel shuffling | Sharpening & Blurring | $IoU_{side\ groove}$ [%] | $IoU_{erosion\ notch}$ [%] | $IoU_{fatigue\ precrack}$ [%] | $IoU_{ductile\ fracture}$ [%] | $IoU_{brittle\ fracture}$ [%] | $IoU_{other}$ [%] | $mIoU$ [%] |
| REF1* | - | + | - | + | - | - | - | 97.3 | 99.4 | 98.0 | 73.0 | 98.5 | - | 94.8 |
| **HET1** | + | + | + | + | - | - | - | **91.0** | **97.8** | **93.3** | **58.4** | **90.6** | **92.5** | **88.8** |
| HET4 | + | + | + | + | + | + | + | 90.2 | 97.9 | 92.2 | 54.5 | 89.7 | 92.3 | 87.7 |
| HOM1 | + | + | + | + | - | - | - | 91.4 | 74.2 | 12.1 | 30.1 | 87.4 | - | 65.2 |
| **HOM4** | + | + | + | + | + | + | + | **96.5** | **99.0** | **96.6** | **64.5** | **97.7** | - | **92.8** |
| **HAR1** | + | + | + | + | - | - | - | **94.3** | **96.5** | **90.4** | **68.8** | **96.1** | **56.6** | **90.0** |
| HAR4 | + | + | + | + | + | + | + | 94.5 | 96.9 | 94.4 | 66.8 | 95.7 | 34.9 | 89.1 |
| **MAX1** | + | + | + | + | - | - | - | **89.4** | **97.4** | **92.0** | **57.9** | **89.2** | **79.3** | **87.6** |
| MAX4 | + | + | + | + | + | + | + | 89.9 | 97.6 | 93.3 | 53.9 | 89.9 | 77.4 | 87.0 |

*Trained fully-supervised.

We found that for our chosen initial learning rate *lr* of 10$^{-3}$ in combination with the simple dataset and augmentation strategy, the model and validation loss (calculated as Dice loss, see Eq. 6) did not decrease anymore after roughly 50 epochs of training, but rather increased again (see Figure 8). This indicates that the initial learning rate was set to high, meaning that the global minima was jumped, and the learning effort stalled. With a smaller learning rate (*lr*=10$^{-5}$) the loss decreased steadily, and no such effect could be observed, but the training duration was increased. This shows, that with different hyperparameter settings and a hyperparameter tuning possibly even better or more efficient models could be trained for the different augmentation settings. Since the hyperparameter tuning was not in the scope of this study, no further analysis on the behavior of HOM1 was made. For all other models trained with a learning rate of 10$^{-3}$ very good prediction qualities were achieved, and HOM4 was considered for the following evaluations.

The result for HOM4 underline an important aspect of the studied training method. With the mIoU being only 2 % smaller than for the REF1 model and surpassing the 90 % threshold, it is shown that the ratio of unlabeled to labeled images $\mu_{u/l}$ of roughly 5 was chosen big enough for the semi-supervised learning to function properly. For our method, the semi-supervised approach only requires roughly 1/6$^{th}$ of the labeling work compared to the supervised approach (32 vs. 200 labeled images). To reinforce this finding, we trained the "MAX" models. For these models all available labeled images from the homogeneous dataset (+176 images and ground truths) were used for the supervised part of the network ($\mu_{u/l}$=1.15), minimizing the unlabeled part to 328 images only. The mIoUs achieved for this ratio of unlabeled to labeled images yielded no increase and were even slightly smaller than for $\mu_{u/l}\approx 5$. Adding more labels from the homogeneous, laboratory setting, was therefore not beneficial as no new features seemed to be learned. Therefore, it can be stated, that not the sheer number of labeled training fracture surface images is key to training success but the sensitive and evenly distributed labeling of fracture surface images of all types and from all domains, as performed for the heterogeneous dataset.

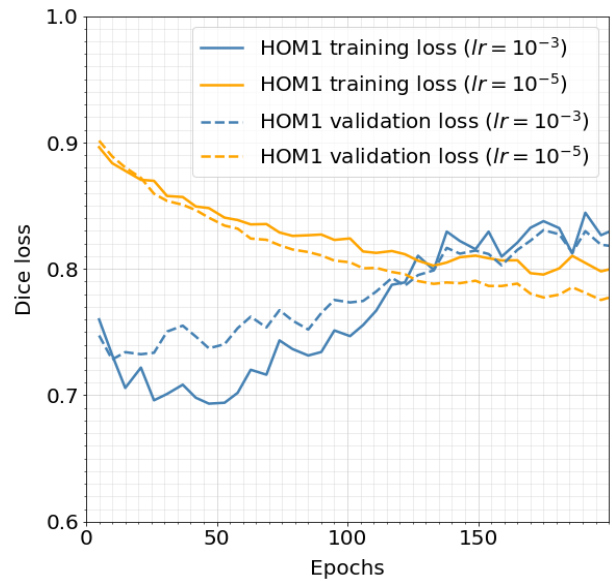

*Figure 8 Epoch-wise development of the model and the validation loss in dependency of the learning rate for HOM1 settings*

For the harmonized dataset models the prediction quality slightly increased for both investigated augmentation strategies, compared to the models trained on the heterogeneous dataset (see Table 4). This shows that increasing the average structural similarity of the training images increases the prediction quality as desired. What both HAR models have in common is the capability of predicting the ductile fracture class better than the HET models, as the IoU was increased by at least 12 %. The main differences in prediction quality between HAR1 and HAR4 lie in the IoUs for the fatigue precrack and the "other" class, as the minimalistic strong augmentation strategy benefitted the recognition of the "other" class, meaning that this model is overall better at distinguishing the class-features. Maximizing the possible perturbations benefitted the fatigue precrack prediction quality (+4 %), while clearly reducing the prediction capability for the "other" class.

Figure 9 shows boxplots for the SSIM values per test dataset and for the mIoU values achieved for the highest scoring model per dataset, enabling a comparison of the distributions between the datasets. The plots show that for each of the three datasets the mIoU boxplots are similar to the SSIM plots. The box sizes, that denote 50 % of the data points, are the smallest for the HOM data, but equally sized in SSIM and mIoU. This is also shown in the SSIM versus mIoU scatter plot below, where all points lie in a small cloud of points. For the HET data, the spread (box and whisker range) for the structural similarity is bigger than for the mIoU values indicating, that the model was able to overcome the difficulties of widespread SSIM, caused by the specimen, material, and imaging-induced differences, to some extent by learning the underlying features. While the SSIM values lie in between 0.06 and 0.5, the mIoU values only scatter from 0.7 to 0.98, which showcases the good generalizability of the model across domains, even though the structural similarity was low. For the HAR data, a behavior like for the homogeneous data and model was found, where the box and whiskers cover a similar range for SSIM and mIoU. Due to the dataset composition, the SSIM versus mIoU cluster is spread wider than for the HOM4 model. This means, the model generalizes on its domain, but is not able to profit that much from the different feature representations as the HET1 model. Overall, the results show there is a high correspondence between the structural similarity and the prediction quality. For homogeneous and harmonized data this behavior was to be expected and the models generalize well for the given data. The results on the heterogeneous dataset show, that the prediction quality may benefit from different images even though their structural similarity is low, as the similarities of the features in the different representations can be learned by the models (scatter of mIoU is smaller than of SSIM) and a very good generalization can be achieved.

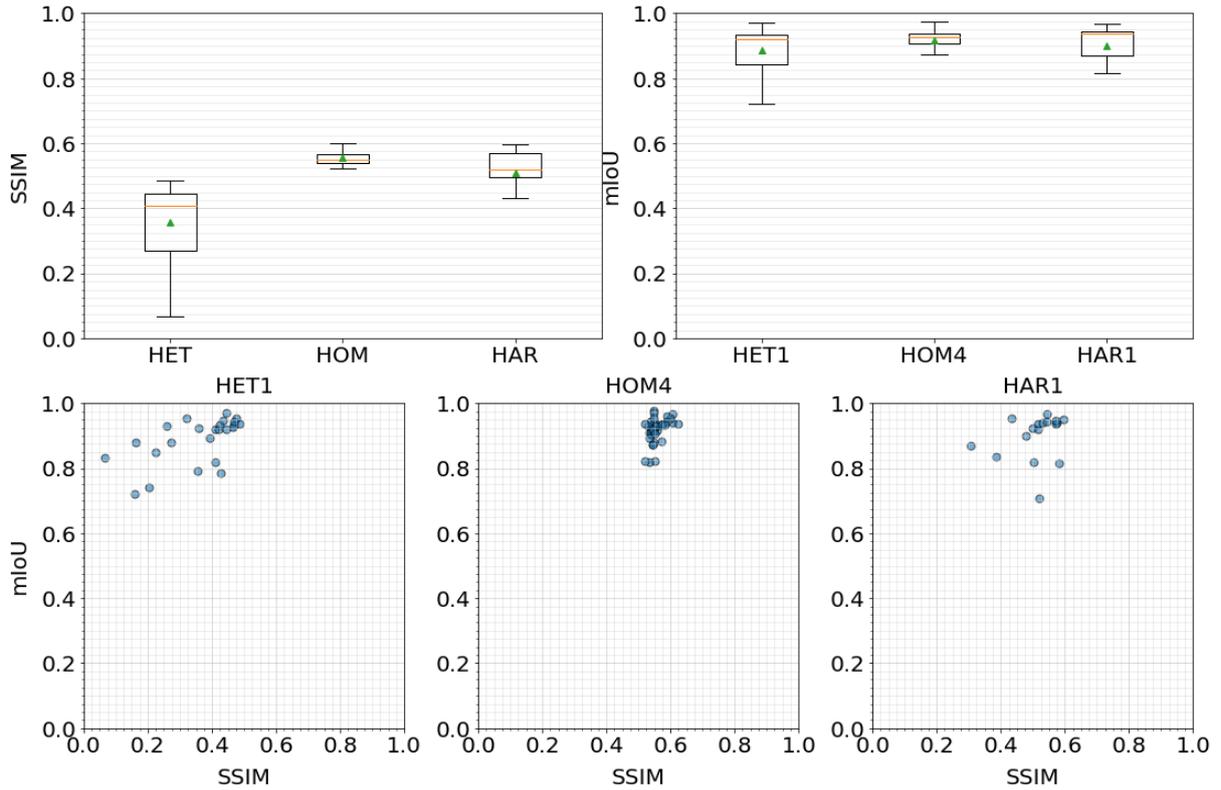

*Figure 9 Top: Boxplots of SSIM distribution for the three datasets and segmentation results for the best model per dataset, Bottom: Segmentation results in dependency of SSIM for the best model per dataset*

### 3.4 Initial crack size measurements

Furthermore, we investigated the capabilities of HET1, HOM4, and HAR1 at initial crack size measurements to demonstrate the benefit for fracture mechanics assessment. For this we used the homogeneous test dataset, where all 38 specimens already had been assessed for their initial crack size by manual 5-point-average measurements (5PA). The automated measurements on the model prediction masks were performed with the area average method (AA). This method is so far not implemented in any ASTM or ISO mechanical test method but the results were found as precise as manual 5-point-average measurements when tested by experts [37, 38] and automated [18].

$$a_0 = \frac{n_{pixels, erosion\ notch} + n_{pixels, fatigue\ precrack}}{B_N} \quad (8)$$

The scatter plots given in Figure 10 show that the automated area average measurements lie in close agreement with the manual 5-point-average measurements for both starter notch lengths respectively crack depth ratios. Only a few measurements fall out of the 1 % measurement deviation scatter band, mostly belonging to the HAR1 model. The biggest deviations are given for the 34th (a), the 27th (b) and 26th (c) test image. For the 26th test image, the HET1 the measurement error is the biggest with an underestimation of 3.729 mm, which is why the point is not shown in Figure 10.

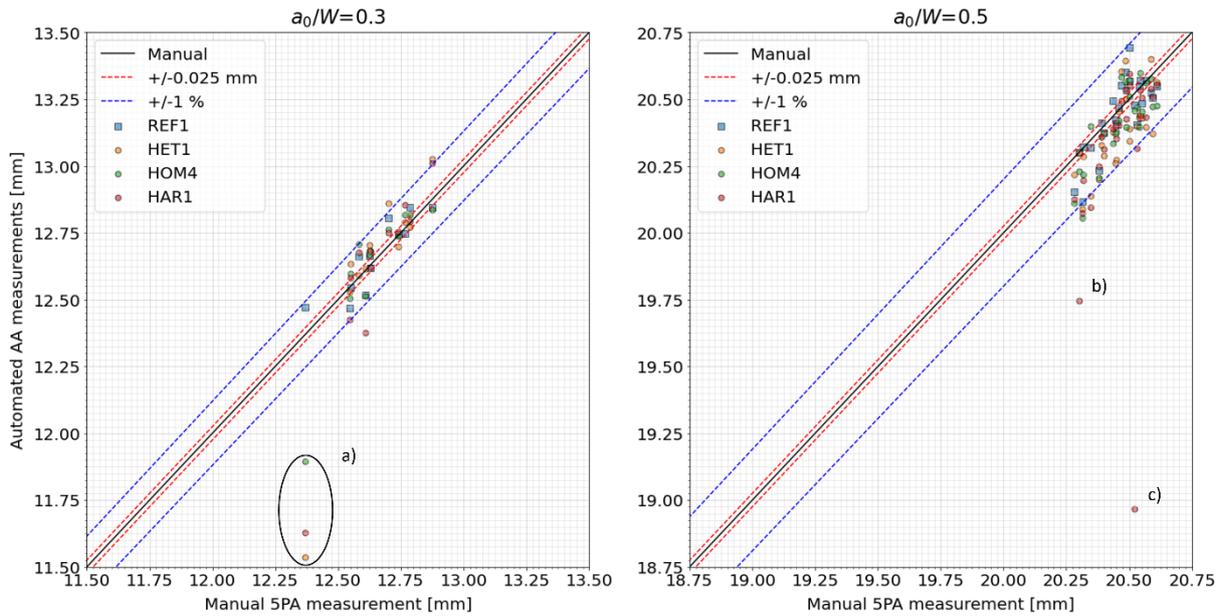

Figure 10 Initial crack size measurement results

To find the causes for the faulty crack size measurements, the prediction masks for the 26th, 27th and 34th test image in Figure 11 can be considered. Misclassified pixels leading to wrong measurement results are indicated by the black circles. For a) a silver staining of the erosion notch can be found in the input image. All models falsely assigned this staining to different classes, leading to a wrong measurement of the "crack area". Input image b) and c) show similar misclassifications for the HAR1 model, where parts of the side grooves are mistaken for fatigue precrack. Due to this the net thickness $B_N$ which directly was measured too long which influences the results of the area average method (see Eq. 8). For c) and HET1 parts of the background (bright scratches visible in the input image) are mistaken for fatigue precrack, resulting in the underestimation of the initial crack size. For the REF1 model no such misclassifications are observable.

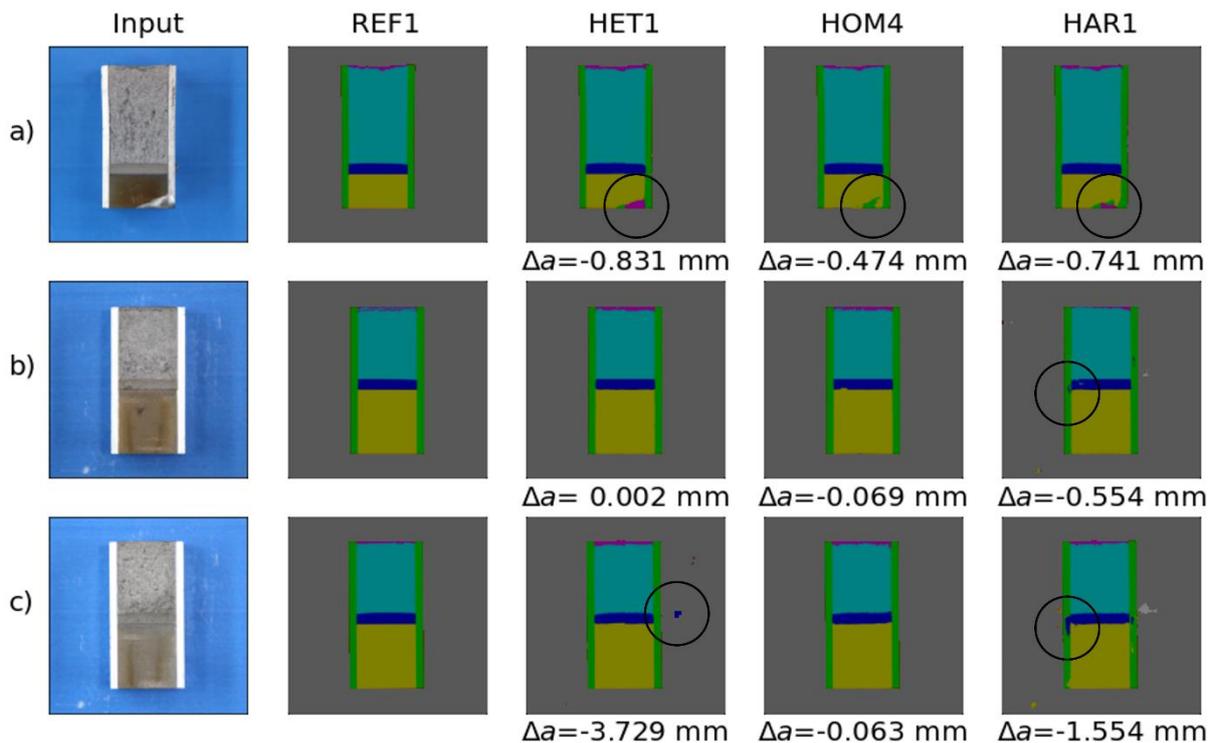

*Figure 11 Comparison of the prediction masks for the crack size measurement outliers*

In Table 6 the absolute and relative initial crack size measurement errors and standard deviations, as well as the prediction quality on the homogeneous dataset (mIoU$_{HOM}$) are given. In accordance with the inferred mIoU values, the REF1 model shows the smallest mean measurement error, while the HET1 model shows the biggest measurement error. Noticeable are the high standard deviations and measurement error for the HET1 and HAR1 model, which can be linked to the outliers described before. Surprisingly the HET1 model outperforms the HAR1 model in prediction quality, while the mean measurement error and standard deviation are lower for HAR1. An important finding here is that the performance of a model in terms of class recognition is therefore not necessarily representative of the accuracy of following tasks such as measurement procedures since they additionally depend on the special arrangements of the erroneous class pixels. If the results for the 26$^{th}$ test image are removed, the results in brackets can be obtained. Now the HET1 and HAR1 model show similar measurement errors and standard deviations. Contrary to the other models, HOM4 learned to generalize for this image, so the results are rarely influenced.

*Table 6 Initial crack size measurement error and standard deviation (values in brackets exclude 26$^{th}$ test image)*

| Model | mIoU$_{HOM}$ [%] | Δa [mm] | σ [mm] | Δa [%] | σ [%] |
|---|---|---|---|---|---|
| REF1 | 94.8 | -0.006 (-0.005) | 0.078 (0.079) | -0.02 (-0.01) | 0.44 (0.45) |
| HET1 | 91.9 | -0.162 (-0.066) | 0.609 (0.167) | -0.82 (-0.35) | 3.10 (1.22) |
| HOM4 | 92.8 | -0.045 (-0.044) | 0.111 (0.112) | -0.24 (-0.24) | 0.76 (0.77) |
| HAR1 | 88.8 | -0.119 (-0.080) | 0.288 (0.167) | -0.64 (-0.45) | 1.62 (1.16) |

Since the measurement results for HET1 are only slightly worse than for HAR1, if the 26$^{th}$ test image is excluded our previous findings could be confirmed. The HET1 model was able to extract and learn common features from the heterogeneous dataset and is now therefore well able to generalize, while being as good at predicting fracture surfaces of SE(B)-40-20 specimens as the HAR1 model, which was trained on curated data. As seen in Table 5, the HAR1 model is better than the HET1 model at distinguishing ductile and brittle fracture, as well as predicting side grooves. However, for classes that are most important for the initial crack size measurement utilizing the arc average method, fatigue precrack, and erosion notch, the HET model is more precise. This is also shown in Figure 12 where the number of pixels per class is once again compared to the prediction qualities for each test image. The mIoU values for HAR1 tend to be on the lower spectrum of the range and especially for the "other" class the resulting mIoU values are higher for the HET1 and HOM4 models. For the HET1 model no misclassifications in images without "other" pixels occur, so that we can assume that the diversity of the real-world training data helped the model learning the different representations of for example gauge notches. Supposedly further increasing the number of images in the heterogeneous and harmonized dataset would therefore lead to an increased measurement precision as the generalization capability would increase. In general, the reference model still shows the highest measurement precision but the especially the HOM4 measurements are of the same quality while requiring much less labeling work.

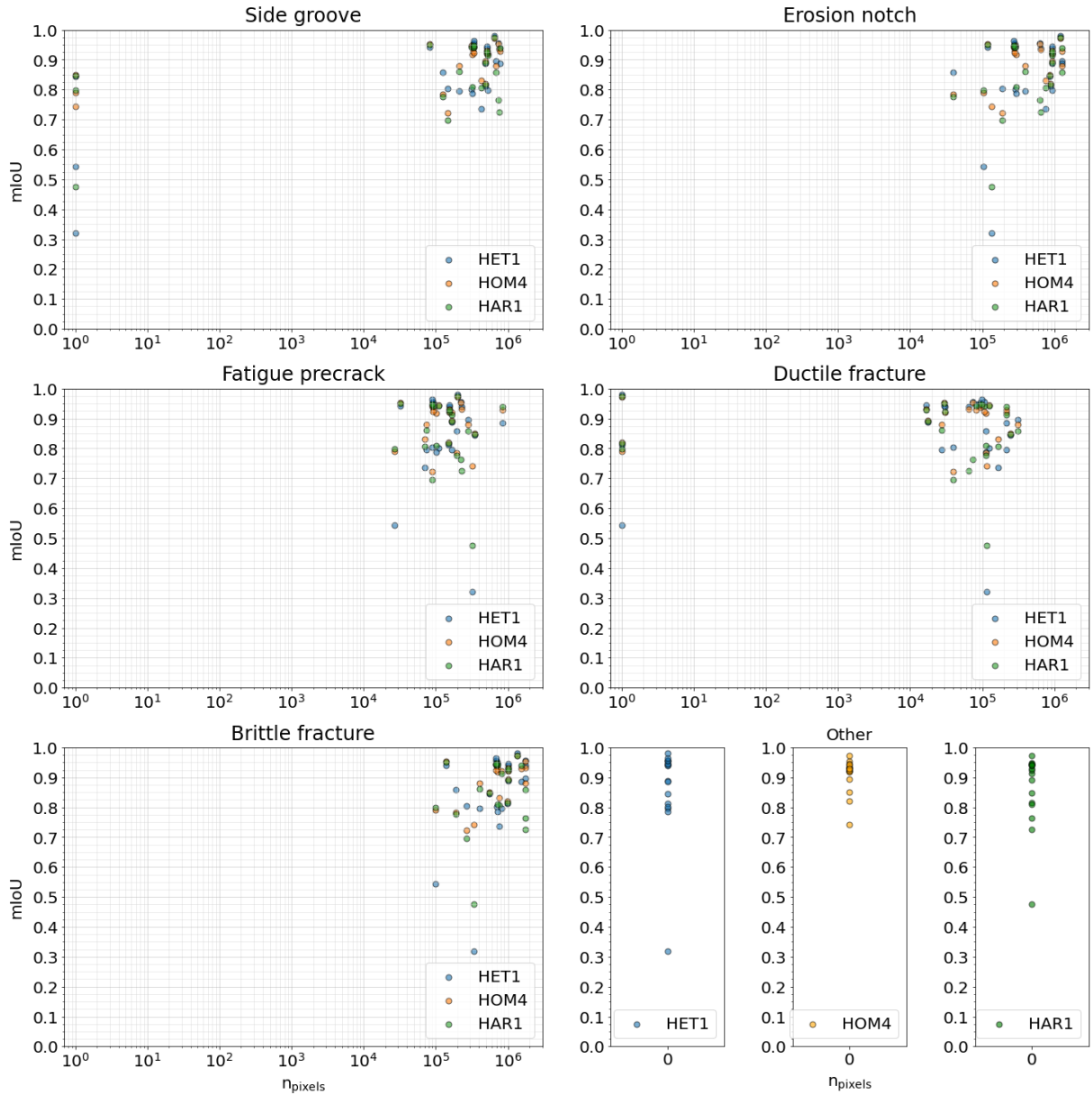

Figure 12 Segmentation results for the homogeneous test dataset in dependence of $n_{pixels}$ for the HET1, HOM4, and HAR1 model ($n_{pixels}=10^0$ denotes images, where the specific class is not present)

# 4. Conclusion

In this study, we showed that it is possible to successfully train semantic segmentation models for macroscale fracture surfaces on real-world datasets with different levels of complexity in terms of their structural similarity. Three image datasets of fracture surfaces with distinct variance were utilized – typical isolated laboratory conditions (homogeneous), aggregated from multiple laboratories (heterogeneous), and a curated subset of the above (harmonized). Following strategies and parameters were found suitable for the weak-to-strong consistency regularized semi-supervised training:

- **Validation data selection:** For the heterogeneous dataset a validation preselection before training (stratified approach) should be performed to avoid possible bias. For homogeneous fracture surface data, a randomized approach is suitable.

- **Semi-supervised learning approach:** Our adapted FixMatch algorithm [28, 32] has shown to greatly benefit the efficiency, since the labeling effort can be significantly reduced. By implementing the adapted weak-to-strong consistency, the amount of labeled data required for model training is reduced by factor 6.
- **Ratio of unlabeled to labeled data:** We found a ratio of unlabeled to labeled data $\mu_{u/l}$ of roughly 5 is suitable for the semi-supervised. A maximization of the labeled training data posed no benefit to the prediction quality of the trained model.
- **Augmentation strategies:** The weak-to-strong consistency required the optimization of both, the weak and the strong, augmentation pipelines. On the one hand, the weak augmentation pipeline benefitted from applying as many non-destructive, and non-rigid transformations as possible. For the strong augmentation pipeline, two approaches utilizing non-spatial transformations showed very good results. For data with a varying structural similarity, e.g., due to imaging-induced variance, keeping the number of perturbations small (e.g., only applying brightness and contrast transformations) showed the best performance. Maximizing the number of possible strong transformations lead to the best prediction quality for the homogeneous data. Augmentations resulting in representations not found in real world scenarios (e.g., gaussian noise) can lower the prediction quality.

Regarding the semi-supervised learning approach, the influence of the structural similarity, and the measurement results, the following findings and outlook could be found:

- **Structural similarity influence:** Models trained on the heterogeneous data are well able to generalize across domains, since they learned feature representations from images covering a wide range of structural similarities, whereas the models trained on the curated and homogeneous data scaled with the structural similarity. Models that were trained on the harmonized data (higher structural similarity) perform better at distinguishing the present classes. E.g., the prediction quality for ductile fracture could be increased by roughly 14 %. However, the models trained on the harmonized dataset are more prone to misclassify pixels as classes that do not occur in the actual images.
- **Measurements:** The measurement error for the initial crack size utilizing the area average method were equally small for models trained on the heterogeneous and the harmonized data. Therefore, we conclude that the performance of a model in terms of class recognition is not necessarily representative of the accuracy of following tasks such as measurement procedures since they additionally depend on the special arrangements of the erroneous class pixels. The deep learning assisted initial crack size measurements had the same quality as manual measurements for the laboratory setting (homogeneous dataset). For models trained on real-world data, for example the heterogeneous dataset, very good measurement accuracies with mean deviations smaller than 1 % could be achieved.
- **Outlook:** For future scopes a hyperparameter-tuning should be considered as for example the initial learning rate can have a big influence on the model training success. Furthermore, the implementation of other semi-supervised learning regularizations and the implementation of ensemble models could be investigated to further increase the prediction quality and measurement performance.

# Acknowledgment

Special thanks to our partners that supported this study with image data and labeling work.
- COMTES FHT, Dobřany, Czech Republic
- George C. Marshall Space Flight Center (NASA), Huntsville, Alabama, USA

- National Institute of Standards and Technology (NIST), Boulder, Colorado, USA
- Oak Ridge National Laboratory (ORNL), Oak Ridge, Tennessee, USA
- SCK CEN, Brussel, Belgium

# Funding

The research project "KEK – Automated analysis of fracture surfaces using artificial neural networks (KNN) for nuclear-relevant safety components" is funded by the German Federal Ministry for the Environment, Nature Conservation, Nuclear Safety and Consumer Protection (Project No. 1501621) on basis of a decision by the German Bundestag.

# Appendix